\documentclass[runningheads]{llncs}
\usepackage[T1]{fontenc}
\usepackage{graphicx}
\usepackage{xcolor}
\usepackage{amsmath}
\usepackage{amssymb}
\usepackage{enumitem}
\usepackage{algorithm}
\usepackage[noend]{algpseudocode}
\usepackage{xspace}
\usepackage{tabularx}
\usepackage{booktabs} 
\usepackage{multirow}
\usepackage{listings}
\usepackage{caption}
\usepackage{subcaption}
\usepackage{hyperref}
\usepackage{wrapfig}

\usepackage{graphicx}
\newcommand{\sayan}[1]{\textcolor{black}{#1}}
\newcommand{\yangge}[1]{\textcolor{black}{#1}}

\newcommand{\srect}{{R}}

\newcommand{\acp}{{\hat{h}}}

\newcommand{\vsys}{{\hat{S}}}
\newcommand{\pc}{{M}}
\newcommand{\asys}{{S_{\pc}}}

\newcommand{\post}{{Reach}}
\newcommand{\apost}{{Reach_{S_\pc}}}
\newcommand{\postfunc}{\textsc{Reach}}
\newcommand{\findpc}{\textsc{DaRePC}}
\newcommand{\compcontr}{\textsc{LearnContract}\xspace}
\newcommand{\refstate}{\textsc{RefineState}}
\newcommand{\refenv}{\textsc{ShrinkEnv}}
\newcommand{\AutoLanding}{{\sf AutoLand}\xspace}
\newcommand{\DroneRacing}{{\sf DroneRace}\xspace}

\newcommand{\posint}{\mathbb{Z}^{\geq 0}}
\newcommand{\sys}{{S}}
\newcommand{\exec}{{\alpha}}

\begin{document}
\title{Refining Perception Contracts: Case Studies in Vision-based Safe Auto-landing
}
%
%
\author{Yangge Li\inst{1}\orcidID{0000-0003-4633-9408} \and
Benjamin C Yang\inst{1}\orcidID{0009-0007-4998-2702} \and
Yixuan Jia\inst{2}\orcidID{0009-0007-5717-5222} \and
Daniel Zhuang\inst{1}\orcidID{0009-0005-6338-5045} \and
Sayan Mitra\inst{1}\orcidID{0000-0002-6672-8470}
}
\authorrunning{F. Author et al.}
%
\institute{
Coordinated Science Laboratory \\ 
University of Illinois Urbana-Champaign\\
\email{\{li213, bcyang2, dzhuang6, mitras\}@illinois.edu} \and 
Massachusetts Institute of Technology\\ 
\email{yixuany@mit.edu}
}
\maketitle              
\begin{abstract}
{\em Perception contracts} provide a method for evaluating safety of control systems that use  machine learning  for perception. A perception contract is a specification for {\em  testing\/} the ML components, and  it gives a method for proving end-to-end system-level safety requirements. The feasibility of contract-based testing and assurance was established  earlier in the context of  straight lane keeping---a 3-dimensional system with relatively simple dynamics.  
This paper presents the analysis of two 6 and 12-dimensional flight control systems that use multi-stage, heterogeneous, ML-enabled perception.  
The paper advances methodology by introducing an algorithm for constructing data and requirement guided refinement of perception contracts (\findpc). 
The resulting analysis provides testable contracts which establish the state and environment conditions under which an aircraft can safety touchdown on the runway and a drone can safely pass through a sequence of  gates. It can also discover   conditions (e.g., low-horizon sun) that can possibly violate the safety of the vision-based control system. 


\end{abstract}

\section{Introduction}
\label{sec:intro}
The  successes of Machine Learning (ML) have led to an emergence of a family of systems that take autonomous control actions based on  recognition and interpretation of signals. 
Vision-based lane keeping in cars and automated landing systems for aircraft are  exemplars of this family. In this paper, we explore two flight control systems that use multi-stage, heterogeneous, ML-enabled perception: an  automated landing system created in close collaboration with our  industry partners and an autonomous racing controller built for a course in  our lab.

\begin{figure}
     \centering
     \begin{subfigure}[b]{0.35\textwidth}
         \centering
         \includegraphics[width=\textwidth]{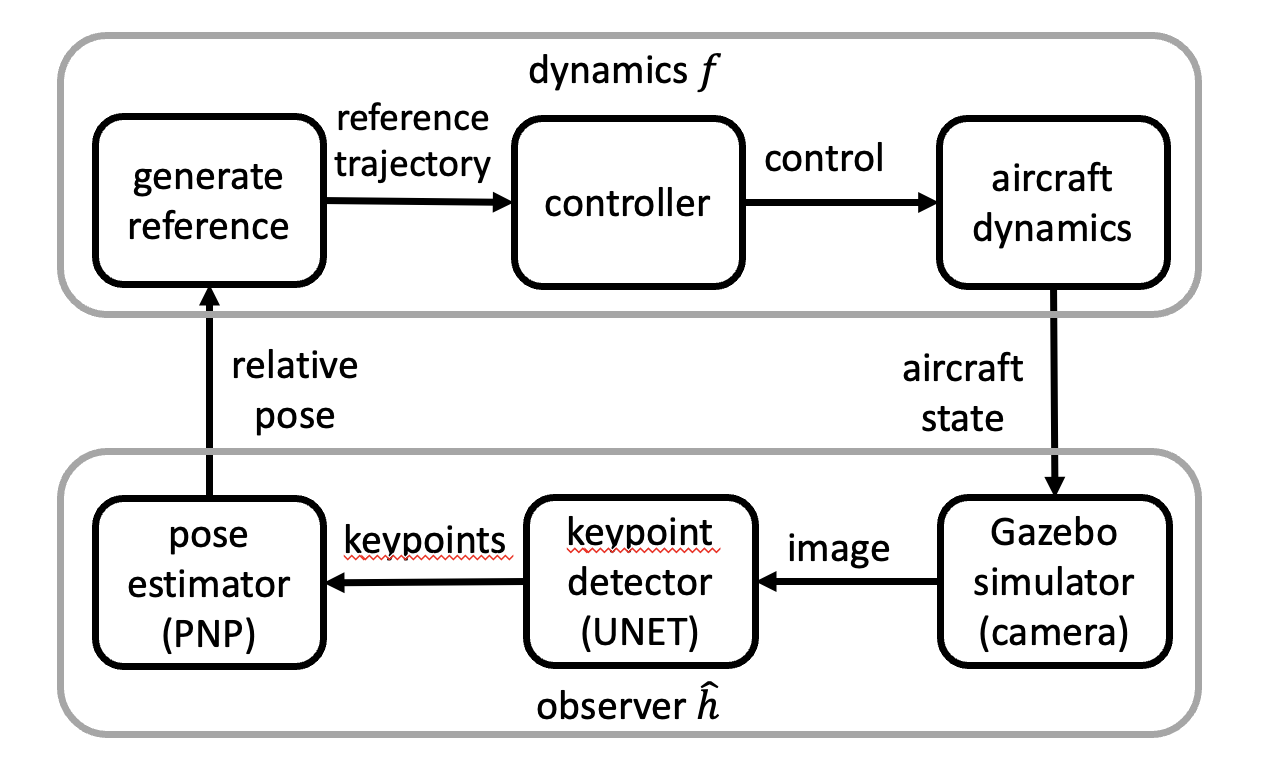}
     \end{subfigure}
     \begin{subfigure}[b]{0.37\textwidth}
         \centering
         \includegraphics[width=\textwidth]{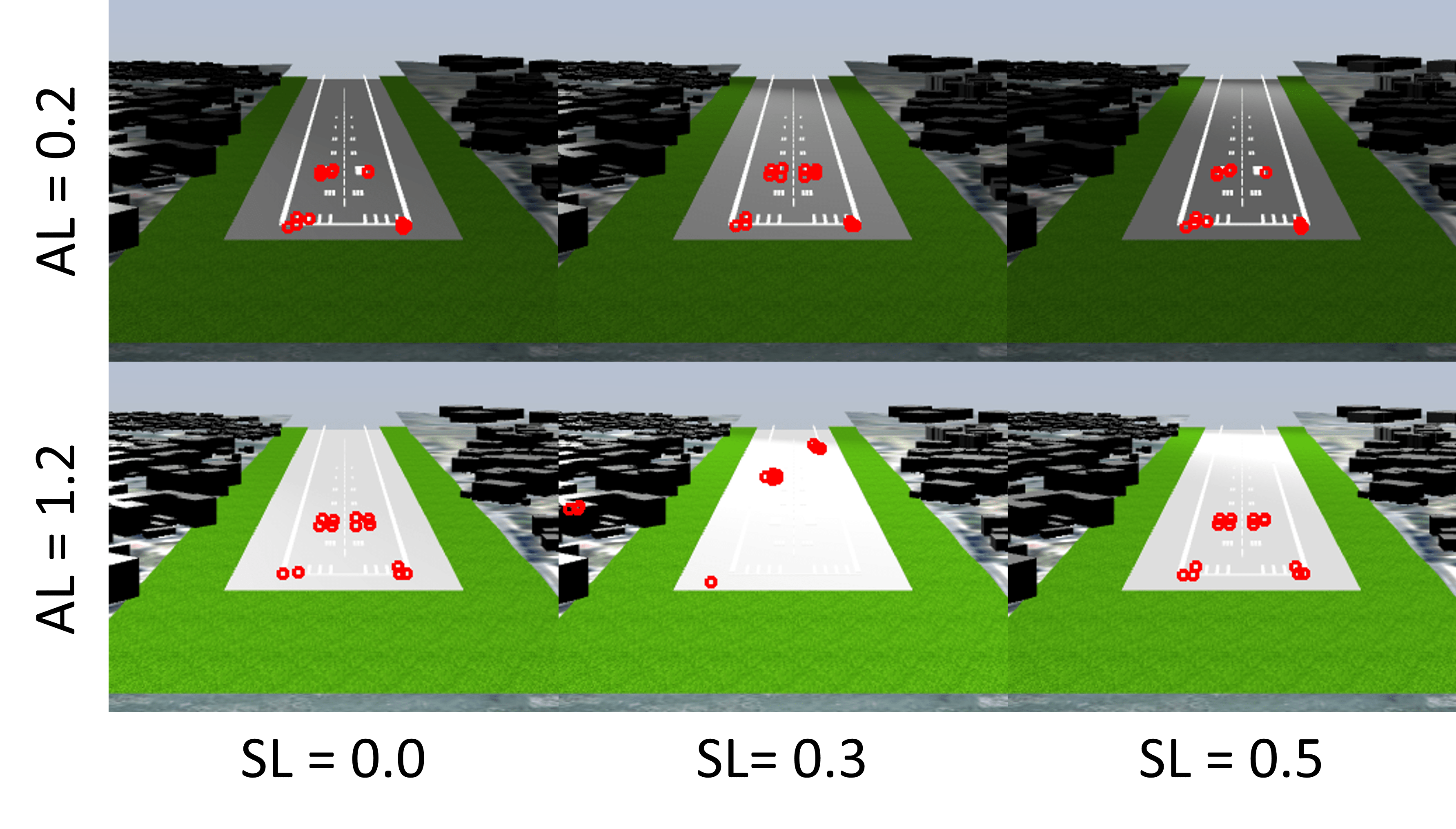}
     \end{subfigure}
     \begin{subfigure}[b]{0.26\textwidth}
         \centering
         \includegraphics[width=\textwidth]{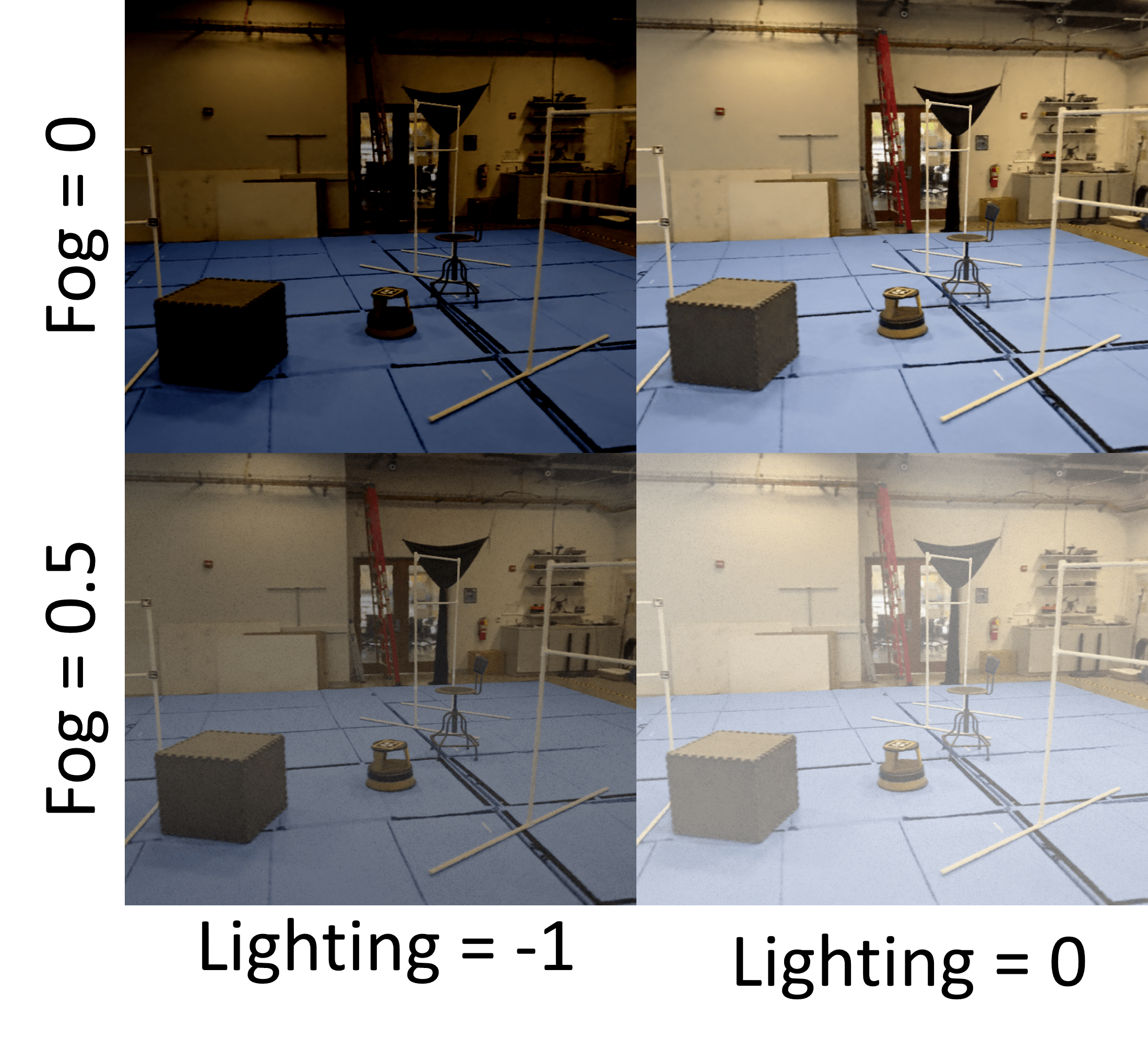}
     \end{subfigure}
    \caption{\small \textit{Left}: Architecture of vision-based automated landing system (\AutoLanding). \textit{Middle}: Camera view and keypoint detection variations with ambient and directed lighting. \textit{Right}: Variation of simulation environment for \DroneRacing.
    }
    \label{fig:exp1:block_diagram}
    \label{fig:exp1:env_params}
    \label{fig:exp2:env_params}
\end{figure}

\paragraph{Vision-based auto-landing system.} 
For autonomous landing, availability of relatively inexpensive cameras and ML algorithms could serve as an attractive alternative. 
The perception pipeline of a prototype vision-based automated landing system (see Fig.~\ref{fig:exp1:block_diagram}) estimates the {\em relative pose of the aircraft with respect to the touchdown area\/} using combination of deep neural networks and classical computer vision algorithm. 
The rest of the auto-landing system uses the estimated pose from the perception pipeline to guide and control the aircraft from 2300 meters---along a reference trajectory of $3^\circ$ slope---down to the touchdown area on the runway.  
The quality of the camera images depend on the intensity and direction of lighting, weather conditions, as well as the actual position of the aircraft and, in  turn, influences the performance of the perception-based pose estimation. From a testing and verification perspective, the problem is to identify a range of environmental conditions and a range of initial approaches from which the aircraft can achieve safe touchdown; safety being  defined by a cone or a sequence of predicates in the state space (see Fig.~\ref{fig:exp1:post_safe}).



\paragraph{Vision-based drone racing.} Drone racing systems are   pushing the boundaries of perception and agile control~\cite{doi:10.1126/scirobotics.adg1462}. Our second case study (\DroneRacing) is about autonomous  control of a drone through a sequence of gates, under different visibility conditions, in a photorealistic simulator we built  using Neural Radiance Fields~\cite{nerfstudio}. The perception pipeline here estimates the position of the drone using a vision-based particle filter.  \sayan{Once again, the challenge is to discover the environment and approach conditions under which the maneuver can be performed reliably.}

\sayan{The problem is challenging because the uncertainties in the perception module are state and environment dependent; these uncertainties have to be  quantified and then propagated through the control and dynamics of the system. }
The core technical challenge for addressing this problem is that we do not have specifications for the DNN-based perception pipeline: What precise property should they satisfy for system-level safety?
The recently introduced    {\em perception contracts} approach~\cite{9852797,angello2023} addresses this problem: The contract can be created from ground truth-labeled data, and it serves a specification for {\em modular testing\/} of the ML perception components. Perception contracts can also be used to {\em prove\/} end-to-end system-level safety requirements. 
%
It gives a bound on the error between true pose $d$ and estimated pose $\hat{d}$ as a function of the ground truth $d$, that is supposed to hold even under environmental variations.
In the previous works~\cite{9852797,angello2023}, the range of environmental factors, and consequently, the perception contracts, were fixed,
and the feasibility of such contract-based reasoning is demonstrated in low-dimensional, and relatively simple applications, such as a lane-keeping system. 
%

This work  advances the perception contracts method in two important ways. First, we develop an algorithm called $\findpc$ that iteratively refines the contracts based both the target system-level requirement and sampled data from the environment. \sayan{To the best of our knowledge,  $\findpc$ is the first approach that can help automatically identify the environmental conditions of vision-based flight control systems,  that can  assure system-level safety.} 
We show that under certain convergence assumptions that can be empirically validated, $\findpc$ terminates with either a contract that proves the requirement or finds a counter-example.
We discuss findings from applying $\findpc$ to \AutoLanding and \DroneRacing. These are challenging applications with high-dimensional (6d and 12d), nonlinear dynamics, and heterogeneous perception pipelines involving Deep Neural Networks and filtering modules. 
For both case studies, we are able to identify a set of environmental conditions where the system is proven to satisfy the requirement. 
\yangge{We show that $\findpc$ is able to identify interesting combinations of non-convex set of environmental conditions that the system can satisfy requirement (Fig.~\ref{fig:exp1:env_safe}).} 
For the \AutoLanding case study, we can further show that our algorithm can find counter-examples violating requirements.

\section{Related Works}
\label{sec:related}

\paragraph{Contracts and Assume-Guarantee.}
A closely related approach has been developed by  P\u{a}s\u{a}reanu, Mangal, Gopinath, and et al. in~\cite{Pasareanu-CAV23,Pasareanu22-controllerRV,Pasareanu2023assumption}. In~\cite{Pasareanu2023assumption}, the authors generate weakest assumptions about the ML component, from the system-level correctness requirements and the available information about the  {\em non-ML} components. These weakest assumptions are analogous to perception contracts, in that they are useful for assume-guarantee style reasoning for proving system-level requirements. However, unlike PCs they are completely agnostic of the actual perception components. The approach was used to analyze a TaxiNet-based lane following system~\cite{Pasareanu-CAV23}. Another important technical difference is that this framework uses  probabilistic discrete state transition systems as the underlying mathematical model.

%
Our previous work~\cite{9852797,angello2023} develops the idea of \emph{perception contracts} using the language of continuous state space models. Thus far, all the applications studies in these threads are related to lane following by a single agent, which is quite different from distributed formation control. \yangge{The idea was recently extended to distributed system to assure safety of vision-based swarm formation control~\cite{hsieh2023assuring}. In \cite{sun2023learningbased}, the perception contract is developed and used to help design a controller for safely landing a drone on a landing pad.}

\paragraph{Safety of automated landing}
Safety of a vision-based automated landing system is analyzed in~\cite{NNlander-NFM22}. This approach carefully exploits the geometry of perspective cameras to create a mathematical relationship between the aircraft pose and perceived pose. This relation is then modeled a NN which is composed with the controller (another NN in this case) for end-to-end analysis. This approach gives strong guarantees at the expense of requiring expertise (in creating the perception model) and being tailored for a particular application. \yangge{Shoukri et al. in \cite{doi:10.2514/6.2021-0998} analyze a vision-based auto landing system using falsification techniques. Their techniques involve computing the range of tolerable error from the perception pipeline and finding states where this error is violated. However, the impacts of environmental variations were not considered in either of these works.}





\paragraph{Perception-based control}
In a series of recent works by Dean et al., the authors studied the robustness guarantees of perception-based control algorithms under simple perception noises: In~\cite{pmlr-v120-dean20a}, the authors proposed a perception-based controller synthesis approach for linear systems and provided a theoretical analysis. In~\cite{pmlr-v155-dean21a}, the authors proposed robust barrier functions that provide an approach for synthesizing safety-critical controllers under uncertainty of the state. More complex perception modules have also been studied. 
In~\cite{9561348}, the reachability of a closed-loop system with a neural controller has been studied.

VerifAI~\cite{verifai-cav19} provides a complete framework for analyzing autonomous systems with ML modules in the loop. \yangge{It is used in \cite{fremont-cav20} to analyze and redesign a Neural Network-Based Aircraft Taxiing System.}


\newcommand{\reals}{\mathbb{R}}

\section{The Problem: Assuring Safety of Vision-based Systems}
\label{sec:setup}

\paragraph{Notations.}
For a function $f: X \rightarrow Y$, we will extend its definition to sets as usual $f(S) := \{ f(x) \ | \ x \in S\}$ for any $S \subseteq X$. Also, for a function defined on sets $f:2^X \rightarrow Y$, we will write $f(\{x\})$ simply as $f(x)$ for $x \in X$.

A vision-based system or simply a {\em system} $S$
is described by a tuple $\langle X, X_0, Y, \\ E, f, o \rangle$ where 
$X \subseteq \reals^n$ is called the {\em state space},
$X_0 \subseteq X$ is the set of initial states,
$Y \subseteq \reals^m$ is called the {\em observation space},
$E$ is called {\em an environment space},
$f: X \times Y \rightarrow 2^X$ is called the {\em dynamics}, and 
$o: X \times E \rightarrow 2^Y$ is called the {\em observer}. 
Typically, $X$ and $f$ are available in analytical form and are amenable to formal analysis. In contrast, the environment $E$ and the observer $o$ are not, and we only have black-box access to $o$. For example, the perception pipelines implementing the observer for the \AutoLanding and the \DroneRacing systems (Sections~\ref{sec:exp1}-\ref{sec:exp2}), are well beyond the capabilities of NN verification tools. 

An {\em execution of a system $S$ in an environment $e \in E$}  is a sequence of states $\alpha = x_0,x_1,\ldots,$ such that $x_0 \in X_0$ and for each $t$, $x_{t+1} \in f(x_t, o(x_t,e)).$ For an execution $\alpha$ and $t \geq 0$, we write the $t^{\mathit{th}}$ state as $\alpha(t)$.
The {\em reachable states of a system $\sys$ in an environment $e\in E$} is a sequence of sets of states $\post_\sys=X_0, X_1,\ldots,$  such that for each $t$, 
$X_t = \cup \exec(t)$ for all possible execution $\exec$. 
For  any $t\geq 0$, $\post_\sys(t)$ denotes the set of states reachable at time $t$. 

A {\em requirement} is given by a set $R \subseteq X$.
%
An execution $\alpha$ in environment $e$ satisfies a requirement $R$ if for each $t\geq 0$, $\alpha(t) \in R$.
The system $S$ satisfies $R$ over $X_0$ and $E$, which we write as $\langle o, X_0, E\rangle \models R$, if for every $e \in E$, every execution of the $S$ with observation function $o$, starting from any state in $X_o$ satisfies $R$. 
\sayan{We write $(o, X_1, E_1) \not\models R$ for some $X_1 \subseteq X$ and $E_1 \subseteq E$, if there exists $x\in X_1, e \in E_1$ and $t\geq 0$, if for the execution $\alpha$ starting from $x$ in $e$, $\alpha(t) \notin R$.}
We make the dependence on $o, X_0,$ and $E$ explicit here because these are the parameters that will change in our discussion.

\paragraph{Problem.}
\sayan{Given a system $S$, we are interested in finding either (a) a subset $E_0 \subseteq E$ of the environment such that $\langle o, X_0, E_0 \rangle \models R$ or (b) a counterexample, that is, an environment $E_c \in E$ such that 
 $\langle o, X_0, E_c \rangle \not\models R$.
}

\section{Method: Perception Contracts}
\label{sec:method}

The system $S$ is a composition of the dynamic function $f$ and the observer $o$, and we are interested in developing compositional or modular solutions to the above problem.  
The observer $o$ in a real system is implemented using multiple machine learning (ML) components, and it depends on the environment in complicated ways. For example, a camera-based pose estimation system of the type used in $\AutoLanding$ uses camera images  which depend on lighting direction, lens flare, albedo and other  factors, in addition to the actual physical position of the camera with respect to the target. Testing such ML-based observers is  challenging because, in general, the observer has no specifications. 

A natural way of tackling this problem is to create an over-approximation $\pc$ of the observer, through sampling or other means. The actual observer $o$ can be made to conform to this over-approximation $\pc$ arbitrarily precisely, by simply making $\pc$ large or conservative. However, such an over-approximation may not be useful for establishing system-level correctness. Thus, $\pc$ has to balance  observer conformance and  system correctness, which  leads to the following notion of {\em perception contract}. 

\begin{definition}
\label{def:pc}
For a given observer $o:X \times E \rightarrow Y$, a requirement $\srect$, and sets $\tilde{X_0} \subseteq \tilde{X} \subseteq X$, $\tilde{E} \subseteq E$, a  $(\tilde{X}_0, \tilde{X}, \tilde{E})$-{\em perception contract\/} is a map  $\pc: X \rightarrow 2^Y$ such that
\begin{enumerate}
    \item Conformance: $\forall x \in \tilde{X}$, $o(x,\tilde{E})\sayan{\subseteq} \pc(x)$
    \item Correctness: $\langle \pc, \tilde{X}_0, \tilde{E}\rangle\models \srect$.
\end{enumerate}
\end{definition}

Recall, $\langle \pc, \tilde{X}_0, \tilde{E}_0\rangle\models \srect$ means that the closed loop system $S_M$, which is the  system $S$ with the observer $o$ substituted by $M$, satisfies the requirement $R$. The following proposition states how perception contracts can be used for verifying actual perception-based systems. 

\begin{proposition}
\label{prop:PCtoR}
Suppose $\tilde{X} \subseteq X$ contains the reachable states of the system  $S = \langle X, \tilde{X}_0, Y, \tilde{E}, f, o \rangle$. 
If  $\pc$ is a $(\tilde{X}_0, \tilde{X}, \tilde{E})$-perception contract for  $o$ for some requirement $R$, then $S$ satisfies $R$, i.e.,  $\langle o, \tilde{X}_0, \tilde{E} \rangle\models \srect$.
\end{proposition}

At a glance,  Proposition~\ref{prop:PCtoR} may seem circular:  it requires us to first compute the reachable states $\tilde{X}$ of $S$ to get a perception contract $M$, which is then used to establish the requirement $R$ for $S$. However, as we shall see with the case studies, the initial $\tilde{X}$ can be a very crude over-approximation of the reachable states of $S$, and the perception contract can be useful for proving much stronger requirements. 



\subsection{Learning Conformant Perception Contracts from Data}
\label{sec:PC-compute}

For this paper, we consider deterministic observers $\acp:X\times E \rightarrow Y$. This covers a broad range of perception pipelines that use neural networks and traditional computer vision algorithms. 
Recall, a perception contract $M:\sayan{\tilde{X}} \rightarrow 2^Y$ gives a set $M(x)$ of perceived or observed values, for any state $x \in \tilde{X}$. In this paper, we choose to represent this set $M(x)$ as a ball $B_{M_r(x)}(M_c(x)) \subseteq Y$ of radius $M_r(x)$ centered at $M_c(x)$. 

In real environments, observers are unlikely to be perfectly conformant to perception contracts.  Instead, we can try to compute a contract $M$ that maximizes some measure of conformance of $\hat{h}$ such as  $p = \mathbb{E}_{x,e \sim  \mathcal{D}}[\mathbb{I}(\hat{h}(x,e) \in M(x))]$, where $\mathbb{I}$ is the indicator function and $\mathcal{D}$ is a distribution over the $X$ and $E$. 
See the case studies and the discussions in Section~\ref{Sec:discussions} for  more information on the choice of  $\mathcal{D}$.
Computing the conformance measure $p$ is infeasible because we only have black box access to $\hat{h}$ and $\mathcal{D}$. Instead, we can estimate an {\em empirical conformance measure:}
Given a set of states $\tilde{X}$ and a set of environments  $\tilde{E}$ over which the perception contract $M$ is computed, we can  sample  according to $\mathcal{D}$ to get $L = \{\langle x_1, e_1, \hat{y}_1 \rangle,...,\langle x_n, e_n, \hat{y}_n \rangle\}$,  where $x_i\in \tilde{X}$, $e_i\in \tilde{E}$, and $\hat{y}_i = \acp(x_i, e_i)$. 
We define the {\em empirical conformance measure\/} of $\pc$ as the fraction of samples that satisfies $\pc$: 
$\hat{p}_n = \frac{1}{n}\sum_{i=1}^{n}\mathbb{I}(\hat{y}_i\in M(x_i))$. 
Using Hoeffding's inequality~\cite{409cf137-dbb5-3eb1-8cfe-0743c3dc925f} 
, we can bound the gap between the actual conformance measure $p$ and the empirical measure $\hat{p}_n$, as a function of the number of samples $n$. 
\begin{proposition}
\label{prop:prob_pc}
    For any $\delta\in (0,1)$, with probability at least $1-\delta$, 
    $p\geq \hat{p} -  \sqrt{-\frac{\ln{\delta}}{2n}}$.
\end{proposition}
%
This gives us a way to choose the number of samples $n$, to bound the gap between the empirical  and the actual conformance, with some probability (as decided by the choice of $\delta$).


\begin{algorithm}
\caption{$\compcontr(\tilde{X}, \tilde{E}, \acp, pr, \epsilon, \delta)$} 
\label{alg:compcontr}
\begin{algorithmic}[1]
    \State $n =- \frac{ln\delta}{2\epsilon^2}$ \label{alg:compcontr:compute_n}
    \State $\{\langle x_i, e_i\rangle\}_{i=1}^n = \textsc{Sample}_{\mathcal{D}}(\tilde{X}, \tilde{E}, n)$ \label{alg:compcontr:samplex} \label{alg:compcontr:samplee}
    \State $\hat{y}_{1}, ... ,\hat{y}_{n} = \acp(x_1, e_{1}), ..., \acp(x_n, e_{n})$ \label{alg:compcontr:samplee}
    \State $L_c = \{\langle x_1, \hat{y}_{1}\rangle,...,\langle x_1, \hat{y}_{n}\rangle\}$ \label{alg:compcontr:cdata}
    \State $M_c = \textsc{Regression}(L_c)$ \label{alg:compcontr:c} 
    \State $L_r = \{\langle x_1, |\hat{y}_{1}-M_c(x_1)|\rangle,...,\langle x_n, |\hat{y}_{n}-M_c(x_n)|\rangle \}$ \label{alg:compcontr:rdata}
    \State $M_r = \textsc{QuantileRegression}(L_r, pr+\epsilon)$ \label{alg:compcontr:r}
    \State \Return $\langle M_c, M_r \rangle$
\end{algorithmic}
\end{algorithm}


The $\compcontr$ algorithm takes as input a set of states $\tilde{X}$, an \sayan{environment set}  $\tilde{E}$ over which a perception contract is to be constructed for the  actual perception function $\acp$. It also takes the desired conformance measure  $pr$, the tolerable gap  $\epsilon$ (between the empirical and actual conformance), and the confidence parameter $\delta$. 
The algorithm first computes the required number of samples for which the gap $\epsilon$ holds with probability more than $1-\delta$ (based on  Proposition~\ref{prop:prob_pc}).
It then samples $n$ states over $\tilde{X}$, and $n$ environmental parameters in $\tilde{E}$, according to a distribution $\mathcal{D}$. The samples  are then used to compute $M_c(x)$ function using standard regression. 
%

The training data for the radius function $M_r$ is obtained by computing the difference between actual perceived state $\hat{h}(x)$ and the center predicted by $M_c(x)$. To make  actual conformance $p\geq pr$, from Proposition~\ref{prop:prob_pc}, we know that the empirical conformance $\hat{p}\geq p+\epsilon \geq pr+\epsilon$.   
Recall,  the empirical conformance is the fraction of samples satisfying  $|\hat{y}_i-M_c(x_i)|\leq M_r(x_i)$. 
$M_r$ is obtained by performing quantile regression with quantile value $pr+\epsilon$. 
Given $L_r = \{\langle x_1, r_1\rangle,...,\langle x_n, r_n\rangle \}$ where $r_i = |\hat{y}_{1}-M_c(x_1)|$, \textsc{QuantileRegression} finds the function $M_r$ such that $pr+\epsilon$ fraction of data points satisfy that $|\hat{y}_i-M_c(x_i)|\leq M_r(x_i)$ by solving the optimization problem~\cite{dekkers2014quantile}:
\[
    \min_{M_r}~\{ (1-\hat{p})\sum_{r_i<M_r(x_i)}{|r_i-M_r(x_i)|}+\hat{p}\sum_{r_i\geq M_r(x_i)}{|r_i-M_r(x_i)|}\}.
\]
Therefore, given $pr$, $\epsilon$, $\delta$, the algorithm is able to find $\pc$ with desired conformance.

\paragraph{Contracts with fixed training data.}
Sometimes  we only have access to a pre-collected training data set. $\compcontr$ can be  modified to accommodate this setup. Given the number of samples $n$ and $pr$, $\delta$, according to Proposition~\ref{prop:prob_pc}, we can get an upper bound on the gap $\epsilon$ and therefore get a lower bound for the empirical conformance $\hat{p}$. We can then use quantile regression to find the  contract with right empirical conformance  on the training data set. 

$\compcontr$ constructs candidate perception contracts that are conformant to a degree,  over specified state $\tilde{X}$ and environment $\tilde{E}$ spaces. For certain environments  $e \in \tilde{E}$, the system
$S$ may not meet the requirements, and therefore,  the most precise and conformant candidate contracts computed over such an environment $e$ will not be correct. In Section~\ref{sec:searchPC}, we will develop techniques for refining $\tilde{X}$ and  $\tilde{E}$ to find contracts that are correct or find counter-examples. For the completeness of this search process we will make the following  assumption about convergence of $\compcontr$ with respect to $\tilde{E}$. 

\paragraph{Convergence of $\compcontr$.} 
Consider a monotonically decreasing sequence of sets $\{E_k\}_{n=1}^\infty$, such that their limit 
$\liminf_{k\rightarrow \infty} E_k = \{e_d\}$ for some environment $e_d \in \tilde{E}$.
Let $\{M_k\}$ be the corresponding sequence of candidate contracts constructed by $\compcontr$ for changing $\tilde{E} = E_k$, while keeping all other parameters fixed. We will say that the algorithm is convergent at $e_d$ if for all $x\in \tilde{X}, \liminf_{k\rightarrow \infty} M_k(x) = \hat{h}(x,e_d).$
Informally, as candidate contracts are computed for a  sequence of environments converging to $e_d$, the contracts point-wise converge to $\hat{h}$. In Section~\ref{sec:exp1}, we will see  examples providing empirical evidence for this convergence property.

\subsection{Requirement Guided Refinement of Perception Contracts}
\label{sec:searchPC}

We  now present {\em Data and Requirement Guided Perception Contract  ($\findpc$)\/} algorithm which uses $\compcontr$ to find perception contract that are correct. 
Consider system $\vsys = \langle \tilde{X}, X_0, Y, \yangge{E_0}, f, \acp\rangle$ and the safety requirement $\srect$, we assume there exists a nominal environment $e_d\in E_0$ such that there exists $x_d\in X_0$ with $\langle \acp, x_d, e_d \rangle\models \srect$.

\begin{algorithm}
\caption{$\findpc(X_0, E_0, e_d, \srect, \acp)$} 
\label{alg:find_e}
\begin{algorithmic}[1]
    \State \textbf{Params}: $\tilde{X}, pr, \epsilon, \delta$ 
    \State $\tilde{E} = E_0$
    \State $\pc_{\tilde{E}}=\compcontr(\tilde{X}, \tilde{E}, \acp, pr, \epsilon, \delta)$ \label{alg:find_e:pc1} 
    \State $queue.push(X_0)$ 
    \While{$queue\neq \emptyset$}
    \State $X_c = queue.pop()$
    \If{$\exists t \text{ s.t } \postfunc(X_c, \pc_{\tilde{E}}, t) \cap \srect = \emptyset$} \label{alg:find_e:cond_unsafe}
    \Return $\langle None, \tilde{E}, X_c\rangle$
    \ElsIf{$\exists t \text{ s.t } \postfunc(X_c, \pc_{\tilde{E}}, t)\nsubseteq \srect$}\label{alg:find_e:cond_unknown}
    \State $X_1, X_2 = \refstate(X_c)$ \label{alg:find_e:refstate}
    \State $queue.push(X_1, X_2)$
    \State $\tilde{E} = \refenv(\tilde{E}, e_d, \pc_{\tilde{E}}, \acp)$ \label{alg:find_e:refenv}
    \State $\pc_{\tilde{E}} = \compcontr(\tilde{X}, \tilde{E}, \hat{h}, pr, \epsilon, \delta)$ \label{alg:find_e:pc2}
    \EndIf
    \EndWhile
    \State \Return $\langle \pc_{\tilde{E}}, \tilde{E}, None \rangle$
\end{algorithmic}
\end{algorithm}

$\findpc$ takes as input initial $X_0$, environment $E_0$ sets, requirement $\srect$, and the  parameters for $\compcontr$.
It outputs \sayan{either} an environment set $\tilde{E}\in E_0$ with a \sayan{candidate contract} $\pc_{\tilde{E}}$ such that $\langle \pc_{\tilde{E}}, X_0, \tilde{E} \rangle\models \srect$ and  the actual observer $\hat{h}$ conforms to $\pc_{\tilde{E}}$ over $\tilde{E}$,
or it outputs a set of initial states $X_c\in X_0$ and an environment set $\tilde{E}\in E_0$ such that $\srect$ will be violated. 


$\findpc$ will check if $\postfunc(P,\pc_{\tilde{E}}, t)$ leaves $\srect(t)$. If $\postfunc(P,\pc_{\tilde{E}}, t)$ completely leaves $\srect(t)$, then $\pc_E$ satisfies $\srect$ can't be found on $X_c$. $X_c$ and $\tilde{E}$ is returned as a counter-example (Line~\ref{alg:find_e:cond_unsafe}).
When $\post(P, \pc_E, t)$ partially exits $\srect(t)$, then refinement will be performed 
by partitioning the initial states using $\refstate$ and shrinking the environments around nominal $e_d$ using $\refenv$ (Line~\ref{alg:find_e:cond_unknown}).
If $\postfunc(P, \pc_E, t)$ never exit $\srect(t)$, then $\pc_E$ is a valid perception contract that satisfy the correctness requirement in Definition \ref{def:pc}. It is returned together with $\tilde{E}$.

We briefly discuss the  subroutines  used in $\findpc$.




\paragraph{$\postfunc(X_c, M_{\tilde{E}}, t)$} computes an over-approximation of $\post_{S_\pc}(t)$ of the system $S_{\pc}=\langle \tilde{X}, X_c, Y, \tilde{E}, f, M_{\tilde{E}} \rangle$. 
For showing completeness of the algorithm, we assume that $\postfunc$ converges to a single execution as $X_c$, $\tilde{E}$ converges to a singleton set. 
This assumption is satisfied by many reachability algorithms, especially simulation-based reachability algorithms~\cite{DryVR,10.1007/978-3-030-99524-9_17}.

\paragraph{$\refstate(X_c)$} takes as input a set of states, and partitions it into a finer set of states. With enough refinements, all partitions will converge to arbitrarily small sets.
\sayan{There are many  strategies for refining state partitions, such as, bisecting hyperrectangular sets along coordinate axes.}

\paragraph{$\refenv(\tilde{E}, e_d, \pc_{\tilde{E}}, \acp)$} The $\refenv$ function takes as input a set of environmental parameters $\tilde{E}$, the nominal environmental parameter $e_d$, $\pc_{\tilde{E}}$ and the actual perception function $\acp$ and generates a new set of environmental parameters. The sequence of sets generated by repeatedly apply $\refenv$ will monotonically converge to a singleton set containing only the nominal environmental parameter $e_d$. 


 \subsection{Properties of $\findpc$}

\begin{theorem}
\label{thm:soundness}
If $\pc_{\tilde{E}}$ is  output from $\findpc$, then $\langle\pc_{\tilde{E}}, X_0, \tilde{E} \rangle \models \srect$
\end{theorem} 
Theorem~\ref{thm:soundness} implies that if the condition in Proposition~\ref{prop:PCtoR} is  satisfied, then $\langle\acp, X_0, \tilde{E} \rangle\models \srect$. 

From the assumptions for $\pc$ and the $\postfunc$ function, we are able to show that as
$\refstate$ and $\refenv$ are called repeatedly, the resulting reachable states computed by $\postfunc$  will converge to a single trajectory under nominal environmental conditions. The property can be stated more formally as: 

\begin{proposition}
\label{prop:ref_convergence}
    Given $S = \langle \tilde{X}, X_0, Y, E_0, f, \acp \rangle$, $\forall x_0\in X_0$ and an execution $\exec$ of $S$ in nominal environment $e_d$, $\forall \epsilon > 0$, $\exists n$ such that $ X_n$ generated by calling $\refstate$ $n$ times on $X_0$ and $\pc_{E_n}$ obtained on environment refined by $\refenv$ $n$ times satisfies that $\forall t$, $\postfunc(X_n, \pc_{E_n}, t)\subseteq B_\epsilon(\exec(t))$  
\end{proposition}

We further define how a system can satisfy/violate the requirement robustly. 

\begin{definition}
    $\vsys=\langle \tilde{X}, X_0, Y, E_0, f, \acp\rangle$ satisfy $\srect$ robustly for $X_0$ and $\tilde{E}$ if $\forall x_0\in X_0$, $\forall e\in \tilde{E}$, the execution $\exec$ satisfies that $\forall t$, $\exists \epsilon>0$ such that $B_\epsilon(\exec(t))\subseteq \srect$.

    Similarly, $\vsys$ violates $\srect$ robustly if there exists an execution $\alpha$ starting from $x_0\in X_0$ under $e_d$ and at time $t$, $\exec(t) \notin \srect$, there always exist $\epsilon>0$ such that $B_\epsilon(\exec(t))\cap \srect = \emptyset$ 
\end{definition}




With all these, we can show completeness of $\findpc$.

\begin{theorem}[Completeness]
\label{thm:completeness}
    If $\pc$ satisfies conformance and convergence and $\vsys=\langle \tilde{X}, X_0, Y, E_0, f, \acp \rangle$ satisfy/violate requirement robustly, then Algorithm \ref{alg:find_e} always terminates, i.e., the algorithm can either find a $\tilde{E}\subseteq E_0$ and $\pc_{\tilde{E}}$ that $\langle  \pc_{\tilde{E}}, X_0, \tilde{E}\rangle\models \srect$ or find a counter-example $X_c\subseteq X_0$ and $\tilde{E}\subseteq E_0$ with $e_d\in \tilde{E}$ that $\langle \acp, X_c, \tilde{E}\rangle\not \models \srect$.   
\end{theorem}

\section{Case Study: Vision-based Auto Landing}
\label{sec:exp1}
Our first case study is $\AutoLanding$, which is a vision-based auto-landing system. Here an aircraft is attempting to touchdown on a runway using a vision-based pose estimator. Such systems are being contemplated for different types of aircraft landing in airports that do not have expensive Instrumented Landing Systems (ILS). The system block diagram is shown in Fig.~\ref{fig:exp1:block_diagram}. The safety requirement $\srect$ defines a sequence of shrinking rectangular regions that the aircraft should remain in (see Fig.~\ref{fig:exp1:post_safe}), otherwise, it has to trigger an expensive missed approach protocol. For our analysis we built a detailed simulation of the aircraft dynamics; the observer or the perception pipeline is the actual code used for vision-based pose estimation. 

\subsection{\AutoLanding control system with DNN-based pose estimator}
The state space $X$ of the \AutoLanding system is defined by the state variables $\langle x,y,z,\psi, \theta, v\rangle$, where $x,y,z$ are position coordinates,  $\psi$ is the yaw, $\theta$ is the pitch and $v$ is the velocity of the aircraft. The perception pipeline implements the observer $\acp$ that attempts to estimate the full state of the aircraft  (relative to the touchdown point), except the velocity. That is, the observation space $Y$ is defined by $\langle \hat{x}, \hat{y}, \hat{z}, \hat{\psi}, \hat{\theta}\rangle$. 

\paragraph{Dynamics.}
The dynamics ($f$) combines the physics model of the aircraft with a controller that follows a nominal trajectory. A 6-dimensional nonlinear differential equation describes the motion of the aircraft with  yaw rate $\beta$,  the pitch rate $\omega$, and  acceleration $a$ as the control inputs (Details are given in the Appendix). The aircraft controller tries to follow a reference trajectory $\xi_r: \posint \rightarrow X$ approaching the runway. 
This reference approach trajectory starts at position $2.3$ kilometers from the touchdown point  and $120$ meters above ground, and has a $-3^\circ$ slope in the z direction. The reference approach has constant speed of $10m/s$. 
%

\paragraph{Vision-based  observer.}
The observations $\langle \hat{x}, \hat{y}, \hat{z}, \hat{\psi}, \hat{\theta}\rangle$ are  computed  using a realistic perception pipeline  composed of  three functions (Fig.~\ref{fig:exp1:block_diagram}): (1) The airplane's front {\em camera}  generates an image, (2) a {\em key-point detector} detects specific key-points on the runway in that image using U-Net~\cite{10.1007/978-3-319-24574-4_28},  
(3) {\em pose estimator} uses the detected key-points and the prior knowledge of the actual positions of those key-points on the ground, to estimate the pose of the aircraft. This last calculation is implemented using the perspective-n-point algorithm~\cite{10.1145/358669.358692}. Apart from the impact of the environment on the image, the observer is deterministic. 


\paragraph{Environment.}
For this paper, we study the effects of two dominant environment factors, namely the ambient light and the position of the sun, represented by a spotlight\footnote{We have also performed similar analysis with fog and rain, but the discussion is not included here owing to limited space.}.
(1) The ambient  light  intensity in the range [0.2, 1.2] 
and (2) the spot light's  angle (with respect to the runway) in the range $[-0.1, 0.6]$ radians, define the environment $E_0$. These ranges are chosen to include a nominal environment $e_d=\langle 1.0, 0\rangle$ in which there exists an initial state $x_0\in X_0$ from which  the system is safe. 
Sample camera images from the same state  under different environmental conditions are shown in Fig.~\ref{fig:exp1:env_params}. 

\subsection{Safety Analysis of \AutoLanding}
\label{sec:analysis:autoland}
We apply $\findpc$ on the $\AutoLanding$ system with reachability analysis  implemented using Verse~\cite{10.1007/978-3-031-37706-8_18}. We run the algorithm with the above range of the environment $E_0$, the requirement $\srect$, and on two rectangular initial sets $X_{01}$ and $X_{02}$ with different range of initial positions (see Appendix~\ref{sec:appendix:init_autoland}).

\begin{figure}[t]
    \centering
    \begin{subfigure}[b]{0.45\linewidth}
        \includegraphics[width=\textwidth, height=5.0cm]{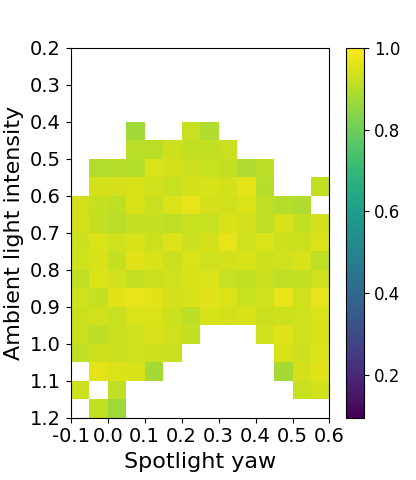}
    \end{subfigure}
    \hfill 
    \begin{subfigure}[b]{0.45\linewidth}
        \includegraphics[width=\textwidth, height=4.5cm]{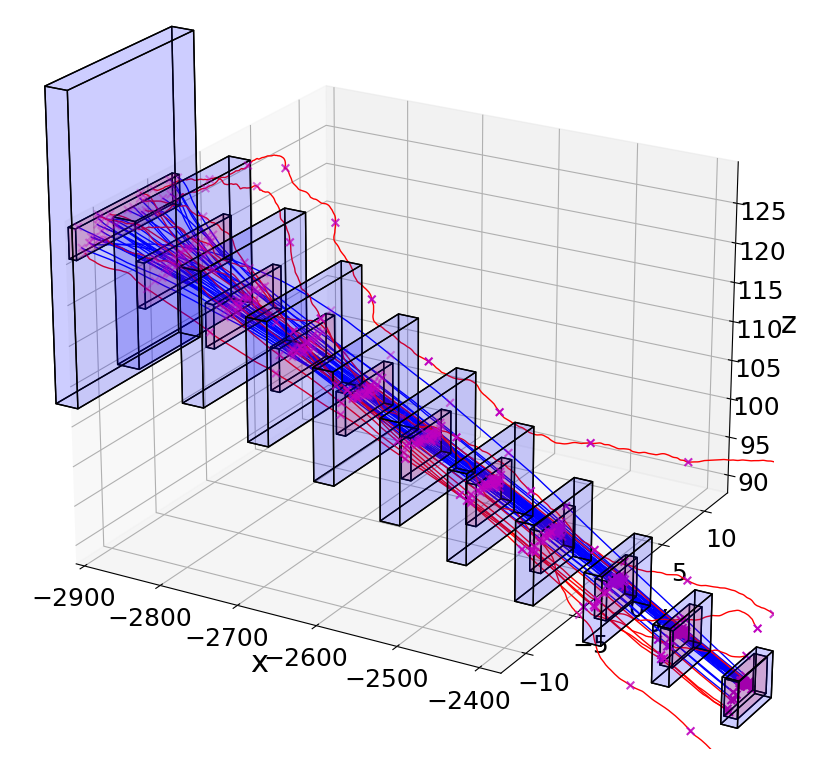}
    \end{subfigure}
    \caption{\small \textit{Left}: Safe environments found by $\findpc$. \textit{Right}:         $\srect$ (blue rectangles), computed $\postfunc$ (red rectangles) and simulations trajectories of vision based landing system starting from $X_{01}$. 
    }
    \label{fig:exp1:env_safe}
    \label{fig:exp1:post_safe}  
\end{figure}

For $X_{01}$, the algorithm finds $\tilde{E}\subset E_0$ and a corresponding perception contract $M_{\tilde{E}}$ such that $\langle \pc_{\tilde{E}}, X_{01},\tilde{E} \rangle \models \srect$ after performing 12 rounds of refinements with empirical conformance measure of $92.4\%$. The resulting $\tilde{E}$ is visualized in Fig.~\ref{fig:exp1:env_safe}. The white region represents the environmental parameters that are removed (not provably safe) during refinement. 
For the colored region,
different colors represent the empirical conformance of $\hat{h}$ to $M$ over each environmental parameter partition. We have following observations from the analysis. 

\paragraph{The computed perception contract $\pc_{\tilde{E}}$ achieves desired conformance.} 
To compute the perception contract, we choose the desired conformance $pr=90\%$, the gap between actual conformance and empirical conformance $\epsilon=1\%$ and the confidence parameter $\delta = 0.1\%$. From $\compcontr$, the empirical conformance measure should be at least $91\%$ for approximately $34$K samples. We round this up to $40$k samples. The computed perception contract $\pc_{\tilde{E}}$ achieves $92.4\%$ empirical conformance measure over $\tilde{E}$. 

Additionally, we run 30 random simulations from $X_{01}$ under $\tilde{E}$. Out of the resulting $30$K data points generated, $29590$ observations were contained in the computed perception contract $M$, which gives an  empirical conformance of $98.6\%$ for this different sampling distribution. 
This shows that $\pc_{\tilde{E}}$ meets the desired conformance and can be even better for some realistic sampling distributions. 

\paragraph{$\findpc$ captures interesting behavior.} The set of environmental parameters $\tilde{E}$ found by $\findpc$ also reveals some interesting behavior of the perception pipeline. We can see from Fig.~\ref{fig:exp1:env_safe} that when the intensity of ambient light is below 0.6 or above 1.15, the performance of perception pipeline becomes bad as many of the partitions are removed in those region. 
However, we can see there is a region $[0.95, 1.1]\times[0.25, 0.4]$ where the ambient lighting condition is close to the nominal ambient lighting condition, but surprisingly, the performance of perception pipeline is still bad. We found that this happens because in this environment, the glare from the spotlight (sun) makes the runway too bright for the  perception pipeline to detect keypoints.
Conversely, in region $[0.4, 0.6]\times [0.25, 0.4]$, the relatively low ambient light is compensated by the sun, and therefore, the perception pipeline has reliable  performance for landing. 

\paragraph{Conformance, correctness and system-level safety.}
Fig.~\ref{fig:exp1:post_safe} shows the system-level safety requirement $\srect$ in blue and $\postfunc(X_{01}, \pc_{\tilde{E}},t)$ in red.
We can see that the computed reachable set indeed falls inside $\srect$, which by  Proposition~\ref{prop:PCtoR}, implies that if the perception pipeline $\hat{h}$ conforms to $\pc_{\tilde{E}}$ then $\langle\acp, X_{01}, \tilde{E}\rangle \models \srect$. 

Further, to check $\langle \acp, X_{01}, \tilde{E}\rangle \models \srect$, we run 30 simulations starting from states randomly sampled from $X_{01}$ under environments randomly sampled from $\tilde{E}$. The results are shown in Fig.~\ref{fig:exp1:post_safe}. All 30 simulations satisfy $\srect$, and only one of these is not in the computed reachable set. This happens because $\pc_{\tilde{E}}$ doesn't have $100\%$ conformance, and therefore cannot capture all behaviors of the actual system. 

\paragraph{Discovering bad environments; $\langle \acp, X_{01}, E_0\textbackslash\tilde{E}\rangle$ doesn't satisfy $\srect$.} 
We performed 10 random simulations from $X_{01}$ under $E_0\setminus \tilde{E}$. 
Eight out of the 10 violate $\srect$ (Red trajectories in Fig.~\ref{fig:exp1:post_safe}). This suggests that the environments removed during refinement can indeed lead to real counter-examples violating $\srect$. 

\begin{figure}[t]
    \centering
    \begin{subfigure}[b]{0.45\linewidth}
        \includegraphics[width=\textwidth,height=5cm]{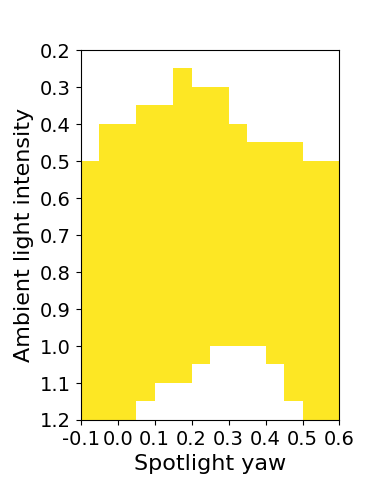}
    \end{subfigure}
    \hfill 
    \begin{subfigure}[b]{0.5\linewidth}
        \includegraphics[width=\textwidth, height=4cm]{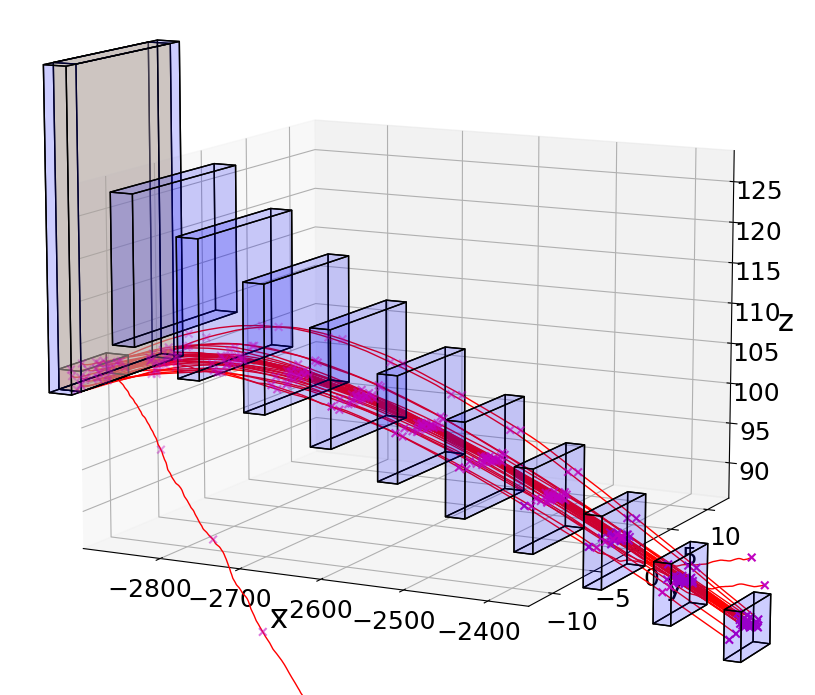}
    \end{subfigure}
    \caption{\small \textit{Left}: Yellow region show $E_c$ found by $\findpc$.\textit{Right}: $\srect$ (blue rectangles) and simulations trajectories (red traces) of vision based landing system starting from $X_c$ (yellow rectangle) under $E_c$. }
    \label{fig:exp1:post_unsafe}  
    \label{fig:exp1:env_unsafe}
\end{figure}

For $X_{02}$, $\findpc$ is able to find a set of states $X_c\subseteq X_{02}$ and $E_c\subseteq E_0$ such that $\langle \acp, x_c, e_c \rangle \not\models \srect$. $X_c$ and $E_c$ are found by calling $\refstate$ on $X_0$ 5 times and $\refenv$ on $E_0$ 8 times. The computed set of environmental parameters $E_c$ and $X_c$ is shown in Fig.~\ref{fig:exp1:post_unsafe}. We perform 30 simulations starting from randomly sampled $x_c\in X_c$ under $e_c\in E_c$ and the results are shown in Fig.~\ref{fig:exp1:post_unsafe}. We can see clearly from the plot that all the simulation trajectories violates $\srect$, which provides evidence that $X_c$ and $E_c$ we found are indeed counter-examples. 


\subsection{Convergence and conformance of Perception Contracts}
\begin{figure}
    \centering
    \begin{subfigure}{0.32\columnwidth}
        \includegraphics[width=\textwidth]{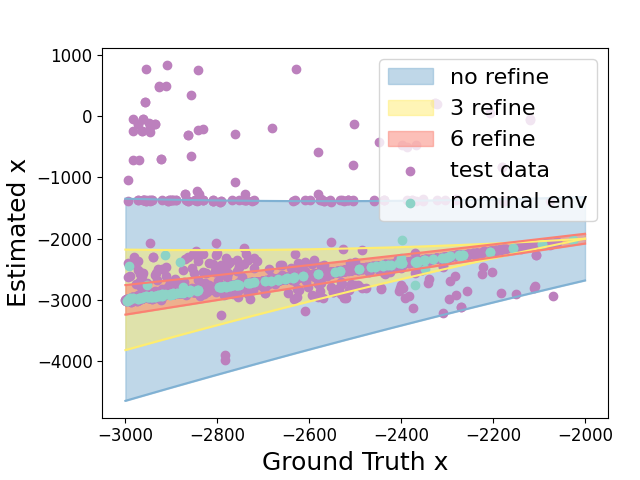}
    \end{subfigure}
    \begin{subfigure}{0.32\columnwidth}
        \includegraphics[width=\textwidth]{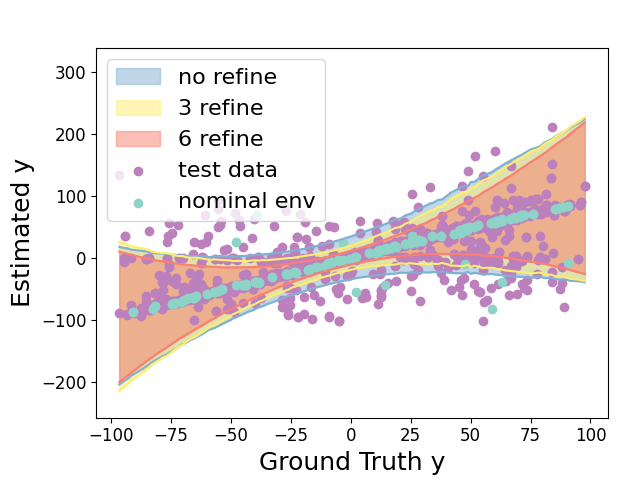}
    \end{subfigure}
    \begin{subfigure}{0.32\columnwidth}
        \includegraphics[width=\textwidth]{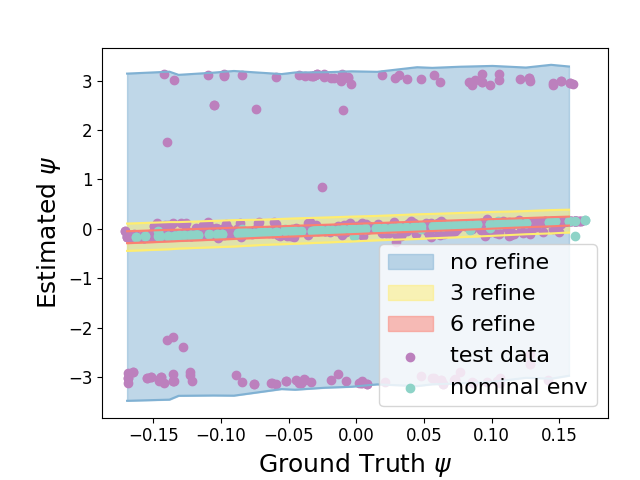}
    \end{subfigure}
    \caption{Visualizing perception contract with different number of refinements for different dimensions. }\label{fig:exp1:pc_all}
\end{figure}

In Section~\ref{sec:PC-compute} , for showing termination of the algorithm, we assume that the perception contract we computed is convergent, i.e. as candidate contracts are computed for a sequence of environments converging to $e_d$, the contracts also  converge to $\hat{h}$ (pointwise). 
To test this, we pre-sampled $80$k data points and set $pr=85\%$, $\delta=0.1\%$. We reduce $\tilde{E}$ for $\pc_{\tilde{E}}$  by invoking $\refenv$ function 0, 3, and 6 times on the set of environments with pre-sampled data points.
$\refenv$ refines the environments by removing environmental parameters where the perceived states is far from the actual states and removing corresponding data points sampled under these environments. 
Fig.~\ref{fig:exp1:pc_all} shows how the output perception contract and its  conformance measure are  influenced by successive refinements. 
%
For the three range of environments $\tilde{E}_0$, $\tilde{E}_3$, $\tilde{E}_6$ after 0,3,6 round of refinements, we observe that $\acp(x, e_d)\in M_{\tilde{E}_6}(x)\subseteq M_{\tilde{E}_3}(x)\subseteq M_{\tilde{E}_0}(x)$, which provides evidence that the perception contract is convergent.  

\paragraph{Refinement preserves conformance measure of perception contract.} As we reduce the range of environmental parameters with 0, 3 and 6 refinements, the training data set reduced from 80000 to 71215 and 62638, respectively. Therefore, according to Proposition~\ref{prop:prob_pc}, the empirical conformance measure should at least be $85.8\%$ for all these three cases to achieve the $85\%$ actual conformance measure. In practice, we are able to get $85.9\%$, $87.5\%$ and $89.1\%$ empirical conformance measure. From this, we can see refinement preserves conformance measure of perception contract.

\section{Case Study: Vision-based Drone Racing (\DroneRacing)}
\label{sec:exp2}
In the second case study, a custom-made quadrotor has to follow a path through a sequence of gates in our laboratory
using a vision-based particle filtering for pose estimation. The system architecture is shown in Fig.~\ref{fig:exp2:block_diagram}. The safety requirement $R$ is a pipe-like region in space passing through the gates. Our analysis here is based on a simulator mimicking the laboratory flying space that we created for this work using Neural Radiance Fields (NeRF)~\cite{nerfstudio}. This simulator enables us to render photo-realistic camera images from any position in the space and with different lighting and visibility (fog) conditions. The observer implementing the vision-based particle filter is the actual code that flies the drone.

\subsection{System Description}
The 12-dimensional state space $X$ of the system  is defined by 
the position $x,y,z$, linear velocities $ vx, vy, vz$, roll pith yaw angles $\phi, \theta, \psi$, and the corresponding angular velocities $\rho, \omega, \beta$.
The observer $\acp$ estimates  the 3d-position and the yaw angle. That is, the observation space $Y$ is defined by $\langle\hat{x}, \hat{y}, \hat{z}, \hat{\psi} \rangle$. \yangge{This is a reasonable setup as roll and pitch angles can be measured directly using onboard accelerometer, but position and yaw angles have to be obtained through computation or expensive external devices such as GPS or Vicon.} 

\paragraph{Dynamics.} A standard 12-dimensional nonlinear model for the drone is used~\cite{8263867}. The input to the aircraft are the desired roll, pitch, yaw angles, and the vertical thrust. The path for passing through the gates is generated by a planner and the controller tries to follow this planned path. More details about the controller and the dynamics are provided in the Appendix~\ref{sec:appendix:drone_dynamics}. 

\paragraph{Observer.} The observations $\langle\hat{x}, \hat{y},\hat{z}, \hat{\psi} \rangle$ are computed using a particle filter algorithm from~\cite{maggio2022locnerf}. 
This is a very different perception pipeline compared to $\AutoLanding$. 
The probability distribution over the state  is represented using a set of particles, from which the mean estimate is calculated.
Roughly, the particle filter updates the particles in two steps: in the {\em prediction step} it uses  the dynamics model to move the particles forward in time, and in the {\em correction step} it re-samples the particles to select  those particles that are  consistent with the actual measurement (images in this case).  The details of this algorithm are not important for our analysis. However, it is important to note that the particles used for prediction and correction do not have access to the environment parameters (e.g. lighting). 

\yangge{
Besides using current image, the particle filter also uses history information to help estimation and therefore the estimation accuracy may improve over time. 
The shape of the reference trajectory that the drone is following can also influence its the performance, especially rapid turning in the reference trajectory. 
Therefore, consider this specific application, we decided to sample data along the reference trajectory for constructing the perception contract which gives similar distribution when the drone is following the reference trajectory. 
}


\paragraph{Environment. } We  study the impact of two environmental parameters: (1) Fog ranging from $0$ (clear) to $1$ (heavy fog), and (2) Ambient light ranging from $-1$ (dark)  and 0 (bright). The nominal environment $e_d=\langle 0,0\rangle$ is a scenario in which the drone is able to complete the path successfully. 
The effect of different environments on the same view are shown in Fig.~\ref{fig:exp2:env_params}. 

\subsection{Analysis of \DroneRacing}
For the \DroneRacing system, the initial set $X_0$ with $x\in [-0.85, -0.83]$, $y\in [-0.06, -0.04]$, $z\in [-0.34, -0.32]$, $\phi\in [-0.01, 0.01]$, $\theta\in [-0.01, 0.01]$, $\psi\in [0.60, 0.62]$ defines an approach zone (all the other dimensions are set to $0$).  
Due to the fact that the perception pipeline for the \DroneRacing example is highly sensitive, we choose desired actual conformance measure $pr = 70\%$, the gap $\epsilon=2\%$ and the confidence parameter $\delta=1\%$. Therefore, the empirical conformance measure should be at least $72\%$ with at least 5756 training data points. 

\begin{figure}[t]
    \centering
    \begin{subfigure}[b]{0.49\linewidth}
        \includegraphics[width=\textwidth,height=2.8cm]{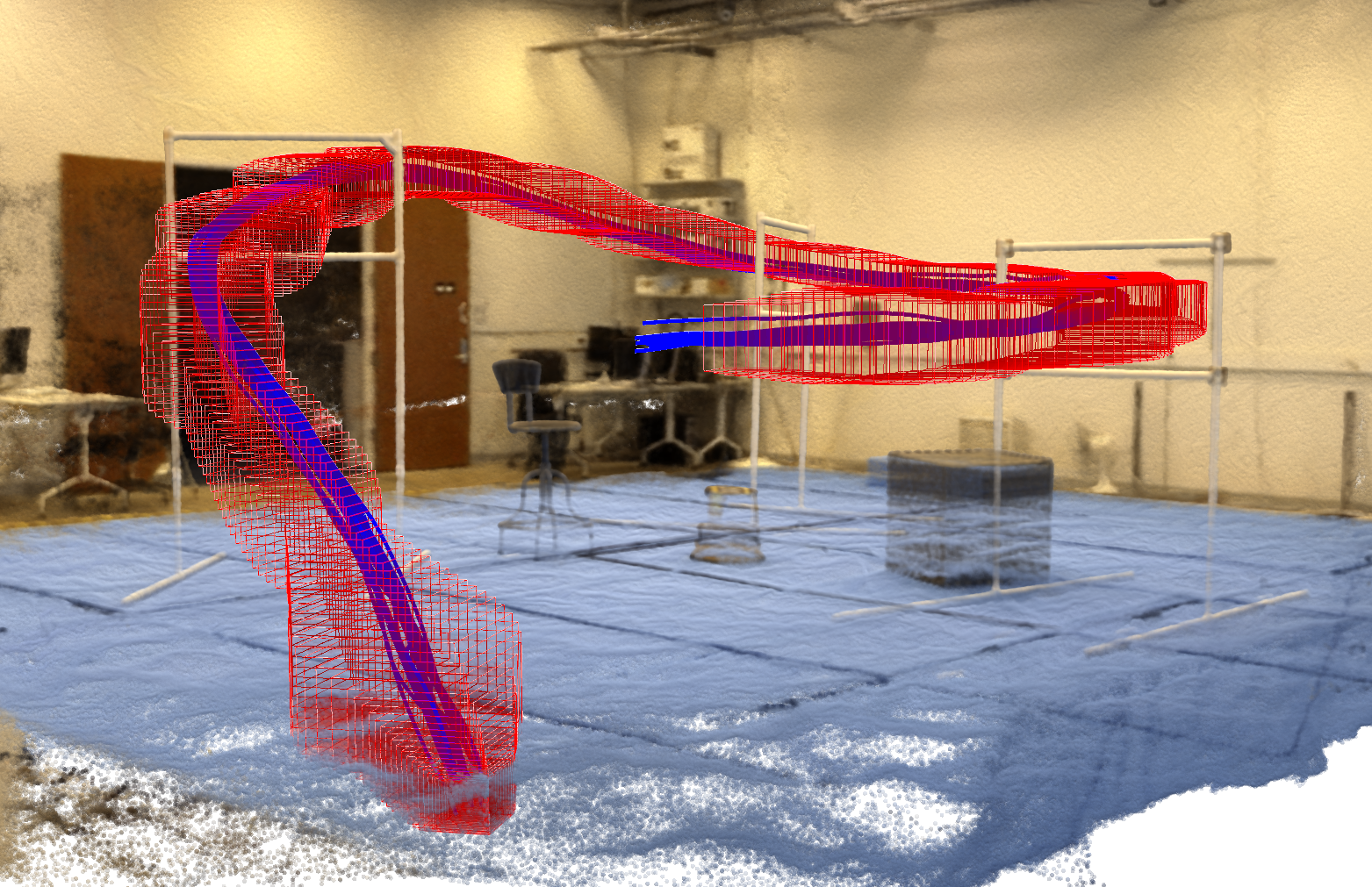}
    \end{subfigure}
    \hfill 
    \begin{subfigure}[b]{0.49\linewidth}
        \includegraphics[width=\textwidth,height=2.8cm]{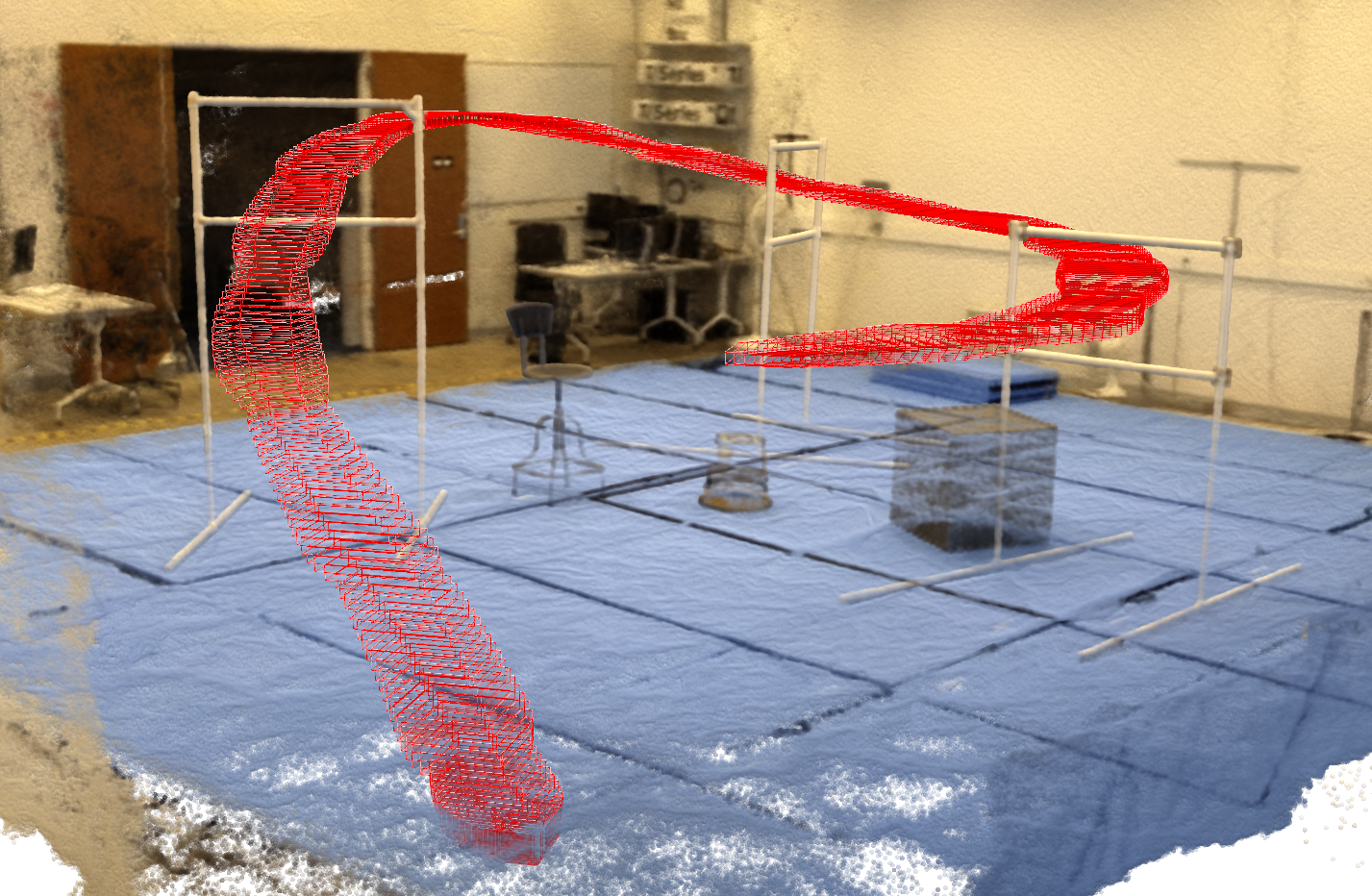}
    \end{subfigure}
    \caption{\small \textit{Left}: Computed $\postfunc$ (red rectangles) and simulations trajectories of \DroneRacing starting from $X_{0}$. \yangge{\textit{Right}: Computed smaller $\postfunc$ (red rectangles) after shrinking $\tilde{E}$ identified by $\findpc$. 
    }}
    \label{fig:exp2:post_safe}  
    \label{fig:exp2:env_safe}
\end{figure}

\paragraph{Conformance, correctness and system-level safety.} 
For $X_0$, with 23 refinements on the environmental, $\findpc$ is able to show that $X_0$ is safe. 
The environment set $\tilde{E}$ that was proved  safe is ball-like region around nominal environment $e_d = \langle 0,0 \rangle$. 
The empirical conformance measure of the  contract is $72.4\%$, which is higher than the minimal empirical conformance needed to get the desired conformance \sayan{with confidence $\delta$}. 
The computed $\postfunc(X_0, M_{\tilde{E}}, t)$ 
reachable set $\srect$ shows correctness with respect to the requirement $\srect$ (Fig.~\ref{fig:exp2:post_safe}), that is,  $\langle \pc_{\tilde{E}}, X_0, \tilde{E} \rangle \models \srect$. 

\paragraph{$\langle\hat{h}, X_0, \tilde{E} \rangle$ satisfies $\srect$.} We run 20 random simulated trajectories starting from randomly sampled $x\in X_0$ and $e\in\tilde{E}$. The results are visualized by the blue trajectories in Fig.~\ref{fig:exp2:post_safe}. We can see all the 20 simulations are covered by the $\postfunc$ computed and therefore doesn't violate $\srect$. This provides evidence that the $\langle \acp, X_0, \tilde{E}\rangle \models \srect$. We also measure the empirical conformance measure of the perception contract for states in these 20 random simulations and the contract is able to get $93.6\%$ empirical conformance measure on these states.

\paragraph{Discovering bad environments.} We run 20 additional simulations starting from randomly sampled $x\in X_0$ under environments randomly sampled from outside of $\tilde{E}$. 
The results are shown in Fig.~\ref{fig:exp2:post_unsafe} in appendix. 
We observe that among all 20 simulations, the drone violates requirments in 19 simulations. This provides evidence that the system can easily violate the requirement outside the environmental factors $\tilde{E}$ $\findpc$ identified. 


\section{Discussions}
\label{Sec:discussions}

There are several limitations of the perception contract method developed in this paper, which also suggest directions for future explorations. 

\paragraph{Distributional Shifts.}
We discussed that $\findpc$ has nice properties of conformance and termination provided $\compcontr$ converges, as the environment converges to the ground conditions. This only makes sense in the limited situation where the  underlying sampling distribution $\mathcal{D}$ for constructing the contract $M$ remains the same under deployed conditions. It will be important to develop techniques for detecting and adapting to distributional shifts in the perception data, in order  to have guarantees that are robust to such changes~\cite{amodei2016concrete}.

\paragraph{Runtime monitoring and control.}
The proplem formulation in this paper is focused on verification, which in turn requires the identification of the environmental conditions under which system safety can be assured. Perception contracts can also be used for runtime monitoring for detection of the violation of certain properties. The key challenge here is to compute  preimages of the contracts, that is: $M^{-1}: Y \rightarrow X$, that can be used for detecting violation or imminent violation of the requirements. Another related problem is the correction of such imminent violations by changing the control actions.

The paper presents the analysis of 6 and 12 dimensional flight control systems that use multi-stage, heterogeneous, ML-enabled perception. Both systems have high-dimensional nonlinear dynamics and realistic perception pipelines including deep neural networks. We explore how the variation of environment impacts system level safety performance. We advances the method of perception contract by introducing an algorithm for constructing data and requirement guided refinement of perception contracts ($\findpc$). $\findpc$ is able to either identify a set of states and environmental factors that the requirement can be satisfied or find a counter example. In future, we are planning to apply the algorithm to real hardware to test its performance. 

\bibliographystyle{splncs04}
\bibliography{egbib,sayan1}
%




\appendix
\newpage
\section{Proof for Proposition~\ref{prop:PCtoR}}
\textbf{Proposition~\ref{prop:PCtoR} Restated~} 
Suppose $\tilde{X} \subseteq X$ contains the reachable states of the system  $S = \langle X, \tilde{X}_0, Y, \tilde{E}, f, o \rangle$. 
If  $\pc$ is a $(\tilde{X}_0, \tilde{X}, \tilde{E})$-perception contract for  $o$ for some requirement $R$, then $S$ satisfies $R$, i.e.,  $\langle o, \tilde{X}_0, \tilde{E} \rangle\models \srect$.
\begin{proof}
    Consider any environment $e \in \tilde{E}$, any initial state $x_0 \in \tilde{X}_0$. 
    From correctness of  $\pc$, it follows that  
    $\langle M, \tilde{X}_0, \tilde{E} \rangle\models \srect$. That is, for the system with $M$ as observer $\asys=\langle X, \tilde{X}_0, Y, \tilde{E}, \pc\rangle$ the reachable sets $\apost(t)\subseteq \srect$ for all $t$, \sayan{and  for each $e \in \tilde{E}$}. 
    Now consider any execution $\alpha$ of $S = \langle X,\tilde{X}_0, Y, \tilde{E}, f, o\rangle$. We know $\alpha(0)\in \tilde{X}_0\subseteq \apost(0)$. 
    Since $\post_S(t)\subseteq \tilde{X}$, from conformance of $\pc$, we know that $o(\alpha(t),\tilde{E})\subseteq \pc(\alpha(t))$ for all $t$. 
    Therefore, for $\alpha(t)\in \apost(t)$, we have $\alpha(t+1) = f(\alpha(t), o(\alpha(t),e))\in f(\alpha(t), M(\alpha(t)))\subseteq \apost(t+1)$. This tells us that $\alpha(t)\in \apost(t)$ for all execution $\alpha$ for all $t$, which gives us $\langle o, \tilde{X}_0, \tilde{E}\rangle \models \srect$.
\end{proof}

\section{Proof for Proposition~\ref{prop:prob_pc}}
\textbf{Proposition~\ref{prop:prob_pc} Restated~}
    For any $\delta\in (0,1)$, with probability at least $1-\delta$, we have that 
    \[
    p\geq \hat{p} -  \sqrt{-\frac{\ln{\delta}}{2n}}
    \]
\begin{proof}
From Hoeffding's inequality, we get 
\[
\begin{split}
Pr(\hat{p} - p\geq \sqrt{-\frac{\ln{\delta}}{2n}}) & \leq e^{-2n(\sqrt{-\frac{\ln{\delta}}{2n}})^2}  = e^{-2n*(-\frac{ln\delta}{2n})} = \delta
\end{split}
\]
This implies that 
\[
Pr(\hat{p}-p\leq \sqrt{-\frac{ln\delta}{2n}}) \geq  1-\delta
\]
which proves the proposition.
\end{proof}

\section{Proof for Proposition~\ref{prop:ref_convergence}}
\textbf{Proposition~\ref{prop:ref_convergence} Restated~}
    Given $S = \langle X, X_0, Y, E_0, f, \acp \rangle$, $\forall x_0\in X_0$ and an execution $\exec$ of $S$ in nominal environment $e_d$, $\forall \epsilon > 0$, $\exists n$ such that $ X_n$ generated by calling $\refstate$ $n$ times on $X_0$ and $\pc_{E_n}$ obtained on environment refined by $\refenv$ $n$ times satisfies that $\forall t$, $\postfunc(X_n, \pc_{E_n}, t)\subseteq B_\epsilon(\exec(t))$ 
\begin{proof}
We know from the definition of $\refenv$ that as the sequence of sets of environmental parameters generated by $\refenv$ will monotonically converge to a singleton set contain the grounding environmental parameter $e_d$. We know from convergence of $\pc$ that $\pc$ will converge to $\acp$ as the set of environmental parameters converge to a singleton set. In addition, we know the output of $\refstate$ will converge to a singleton set of states after it's repeatedly invoked by definition. Therefore, by the assumption of $\postfunc$, when a partition $X$ converges to some $x\in X$ and $\tilde{E} \rightarrow \{e_d\}$, $\postfunc(X, \pc_{\tilde{E}}, t) \rightarrow \exec(t)$ which is the trajectory starting from state $x$ generated by $\vsys$ under environmental parameter $e_d$.   
\end{proof}

\section{Proof for Theorem~\ref{thm:completeness}}
\textbf{Theorem~\ref{thm:completeness} Restated~}
    If $\pc$ satisfies conformance and convergence and $\vsys=\langle X, X_0, Y, E_0, f, \acp \rangle$ satisfy/violate requirement robustly, then Algorithm \ref{alg:find_e} always terminates, i.e., the algorithm can either find a $\tilde{E}\subseteq E_0$ and $\pc_{\tilde{E}}$ that $\langle  \pc_{\tilde{E}}, X_0, \tilde{E}\rangle\models \srect$ or find a counter-example $X_c\subseteq X_0$ and $\tilde{E}\subseteq E_0$ with $e_d\in \tilde{E}$ that $\langle \acp, X_c, \tilde{E}\rangle\not \models \srect$. 
\begin{proof}


To show completeness of the algorithm, we need to show four cases:

1) When there exist a counter-example, the algorithm can find the  counter-example. If there exist a counter-example such that trajectory $\exec_c$ starting from $x_c\in X_0$ under grounding environmental parameter $e_d$ exits $\srect$ at time $t$, then according to the robustness assumption of the system, there exist $\epsilon >0$, such that $\forall x \in B_{\epsilon}(\exec_c(t))$, $x\notin \srect$. Therefore, according to Proposition \ref{prop:ref_convergence}, with enough refinement, we can get a set of states $X_c\subset X_0$ with $x_c\in X_c$ and $\tilde{E}\subset E_0$ with $e_d\in \tilde{E}$ such that $\exec_c(T)\in \postfunc(X_c, \pc_{\tilde{E}}, t)\subseteq B_\epsilon(\exec_c(T))$. In this case, $\postfunc(P, \pc_{\tilde{E}}, t)\cap \srect=\emptyset$ which captures the counter-example. 

2) A counter-example returned is indeed a counter-example. A set of states $X_c$ is returned as counter-example when $\exists t$, such that $\postfunc(X_c, \pc_{\tilde{E}}, t)\cap \srect=\emptyset$ for some $\pc_{\tilde{E}}$. 
From conformance of $\pc_{\tilde{E}}$, we know $\acp(x, e)\in M(x)$ for all $x$ and $e\in \tilde{E}$
In this case, trajectory $\exec$ starting from $x_c\in X_c$ with any environmental parameter $e\in \tilde{E}$ will satisfy that $\exec(t)\in \post(P,\pc_{\tilde{E}},t)$ which is not in $\srect$. The output of $\findpc$ in this case is indeed a counter-example. 

3) When a counter-example doesn't exist, the algorithm can find $\pc_{\tilde{E}}$ satisfy $\srect$ on $X_0$. This case can be shown similar to case 1). Since we assume the system is robust, for every trajectory of the system that satisfies $\srect$, there exist $\epsilon>0$ such that $B_\epsilon(\exec(t))\subset \srect$ for all $t$. Since we know as $\findpc$ repeatedly performs refinement, $\postfunc(X, \pc, t)$ will converge to a single execution. Therefore, according to Proposition \ref{prop:ref_convergence}, with enough refinement, for all safe trajectory we can get $\exec(t)\in \postfunc(X,\pc,t)\subset B_\epsilon(\exec(t))\subset \srect(t)$. Therefore, with enough refinement, we can always find a $\pc_{\tilde{E}}$ that satisfies $\srect$ when the $\vsys$ satisfy $\srect$ on $X_0$ under $e_d$.

4) A $\pc_E$ returned by $\findpc$ will always satisfy $\srect$ on $X_0$. This is given automatically by Theorem \ref{thm:soundness} and the conformance assumption. 
\end{proof}

\section{Dynamics and Controller for \AutoLanding Case Study}
The dynamics of the \AutoLanding Case Study is given by 
\[
\begin{split}
& x_{t+1} = x_t + v_tcos(\psi_t)cos(\theta_t)\delta t\\
& y_{t+1} = y_t + v_tsin(\psi_t)cos(\theta_t)\delta t\\
& z_{t+1} = z_t + sin(\theta_t)\delta t\\ 
& \psi_{t+1} = \psi_t + \beta\delta t\\ 
& \theta_{t+1} = \theta_t + \omega\delta t\\ 
& v_{t+1} = \theta_t + a\delta t
\end{split}
\]
$\delta t$ is a descretization parameter. The control inputs are the yaw rate $\beta$, pitch rate $\omega$ and acceleration $a$

The error between the reference points and state is then given by 
\[
\begin{split}
& x_e = cos(\psi)(x_r-x)+sin(\psi)(y_r-y)\\
& y_e = -sin(\psi)(x_r-x) + cos(\psi)(y_r-y)\\ 
& z_e = z_r-z\\  
& \psi_e = \psi_r - \psi\\  
& \theta_e = \theta_r-\theta\\  
& v_e = v_r-v
\end{split}
\]

The control inputs are then computed using errors computed by a PD controller. 
\[
\begin{split}
& a = 0.005(\sqrt{(v_rcos(\theta_r)cos(\psi_r)+0.01x_e)^2+(v_rsin(\theta_r)+0.01z_e)^2} - v) \\ 
& \beta = \psi_e + v_r(0.01y_e+0.01sin(\psi_e)\\ 
& \omega = (\theta_r-\theta)+0.001z_e
\end{split}
\]

\section{Initial Sets for \AutoLanding Case Study}
\label{sec:appendix:init_autoland}
The initial sets $X_{01}$, $X_{02}$ for \AutoLanding case study are 
\begin{align*}
\small
X_{01} = [-3020, -3010] \times [-5,5] \times [118,122] \times [-0.001, 0.001] \times [-0.0534, -0.0514] \times [9.99, 10.01] \\ 
X_{02} = [-3030,-3000]\times[-5,5]\times[100,140]\times [-0.001, 0.001] \times [-0.0534, -0.0514] \times [9.99, 10.01]. 
\end{align*}

\section{Block Diagram for \DroneRacing}
\begin{figure}[t]
    \centering
    \includegraphics[width=\textwidth]{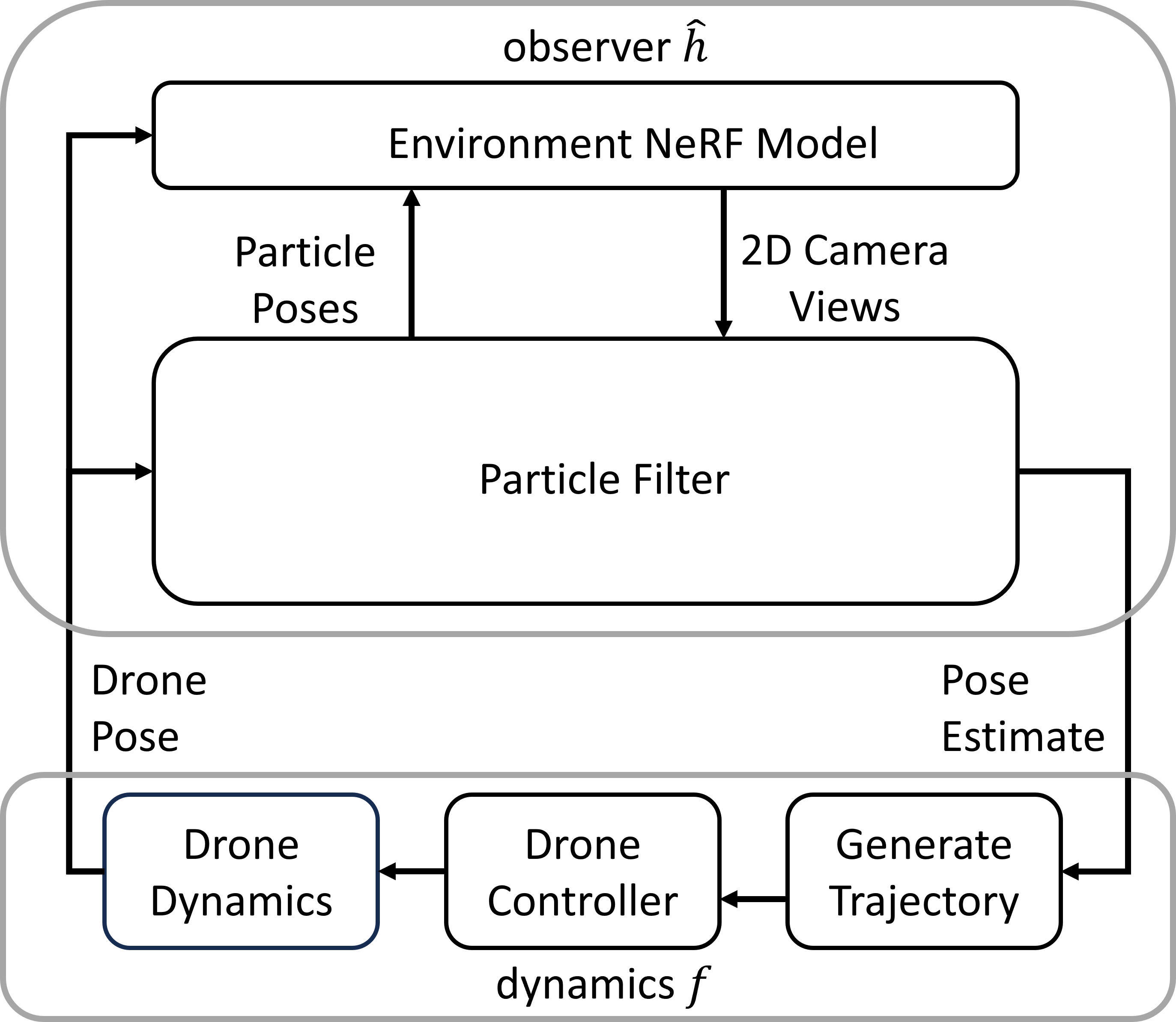}
    \caption{\small Architecture of drone racing system (\DroneRacing).}
    \label{fig:exp2:block_diagram}
    \hfill 
\end{figure}

The block diagram for the \DroneRacing case study is shown in Fig.~\ref{fig:exp2:block_diagram}

\section{Dynamics and Controller for \DroneRacing Case Study}
\label{sec:appendix:drone_dynamics}
We used a slightly modified version of drone dynamics from \cite{8263867}. It's given by the following differential equations
\[
\begin{split}
    & \dot{x} = vx *cos(\psi)-vy *sin(\psi)\\
    & \dot{vx} = g *tan(\phi)\\
    & \dot{\phi} = -d1 *\phi+\rho\\
    & \dot{\rho} = -d0 *\phi + n0*ax\\
    & \dot{y} = vx *sin(\psi)+ vy *cos(\psi)\\
    & \dot{vy} = g *tan(\theta)\\
    & \dot{\theta} = -d1 * \theta + \omega\\
    & \dot{\omega} = -d0 * \theta + n0 * ay\\
    & \dot{z} = vz\\
    & \dot{vz} = kT*F-g\\
    & \dot{\psi} = \beta \\
    & \dot{\beta} = n0*az
\end{split}
\]
Where some parameters $g=9.81$, $d0 = 10$, $d1 = 8$, $n0 = 10$, $kT = 0.91$. Given reference state provided by the reference trajectory, we can compute the tracking error between reference state and current state $err=[x_e, vx_e, \phi_e, \rho_e, y_e, vy_e, \theta_e, \omega_e, z_e, vz_e, \psi_e, \beta_e]$. The position error between current state of the drone and the reference trajectory is given by.
\[
\begin{split}
    & \begin{bmatrix}
        x_e\\y_e\\z_e 
    \end{bmatrix} = 
    \begin{bmatrix}
        cos(\psi)& -sin(\psi)& 0 \\ sin(\psi)& cos(\psi)& 0\\ 0 & 0 & 1 
    \end{bmatrix} 
\end{split}
\]
The tracking error for other states is given by the difference between reference state and current state. 
Therefore, the control signals $ax, ay, kT, az$ can be computed using following feedback controller. 
\[
\begin{split}
    & \begin{bmatrix}
        ax\\ay\\kT
    \end{bmatrix} = 
    \begin{bmatrix}
        3.16  & 4.52  & 4.25  & 1.36  & 0.   & -0.   & -0.    & 0.   & -0.   &  0.  \\
        0.    & 0.    & 0.    & 0.    & 1.   &  1.83 &  1.46  & 1.14 & -0.   & -0.  \\
        0.    & 0.   & -0.    & 0.    & 0.   & -0.   & -0.   & -0.   &  1.   &  1.79
    \end{bmatrix} err \\ 
    & az = 1.0*\psi_e + 1.0*\beta_e
\end{split}
\]

\section{Simulation Trajectories for $e\notin \tilde{E}$ for \DroneRacing}
\begin{figure}
    \centering
    \includegraphics[width=\textwidth]{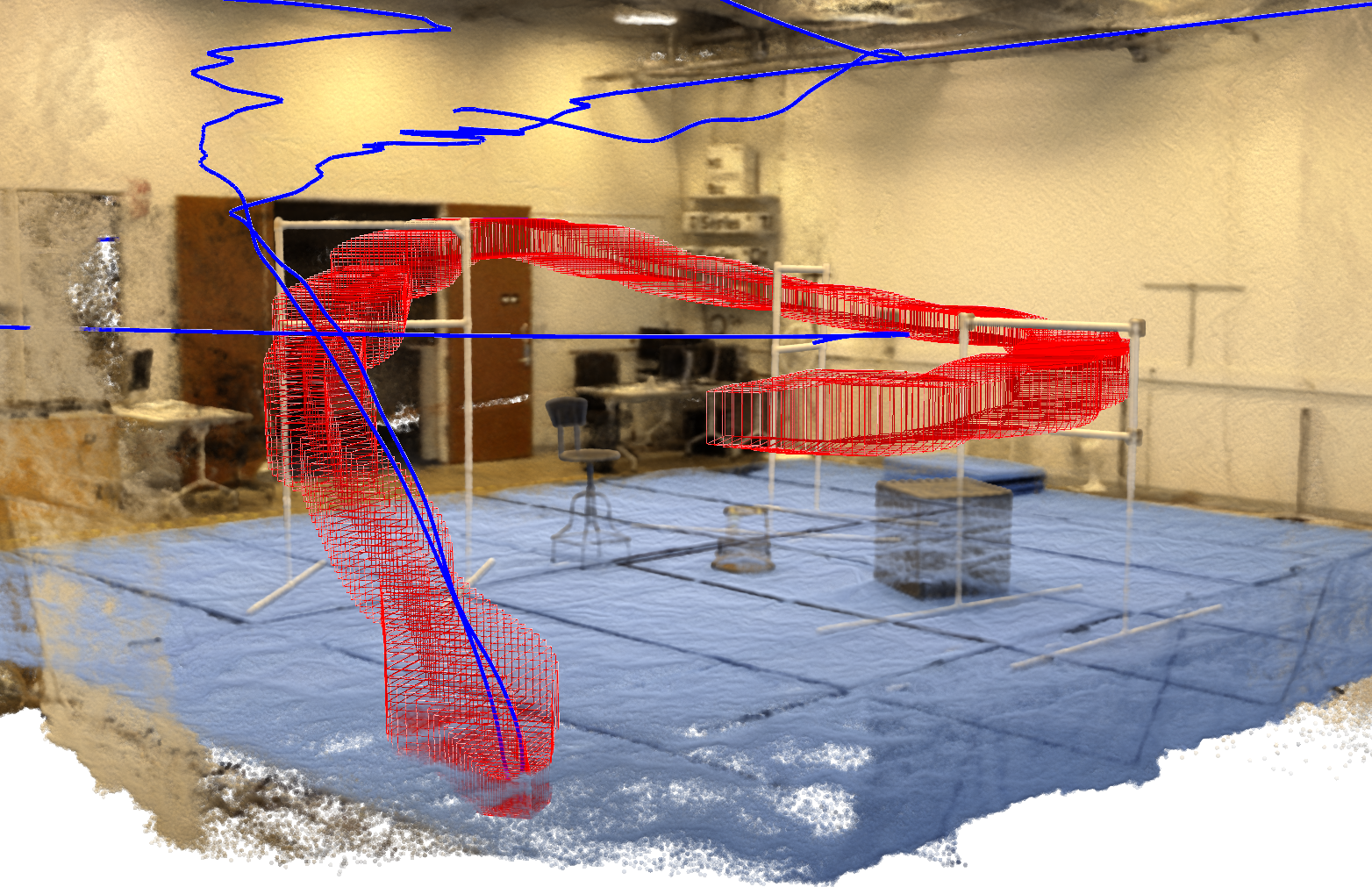}
    \caption{Simulation trajectories under environments $e\in E_0\textbackslash \tilde{E}$ violates requirements. 
    }
    \label{fig:exp2:post_unsafe}
\end{figure}

\end{document}